\documentclass[conference,a4paper]{IEEEtran}
\IEEEoverridecommandlockouts

\usepackage[hidelinks]{hyperref}
\usepackage[cmex10]{amsmath}
\usepackage{amssymb,amsfonts}
\usepackage{dblfloatfix}

\usepackage[ruled,vlined]{algorithm2e}
\usepackage{graphicx}

\usepackage{booktabs}
\usepackage{siunitx}
\usepackage[square,sort&compress,numbers]{natbib}
\usepackage{texnames}
\usepackage{bm,bbm}
\usepackage{orcidlink}
\usepackage{multirow}
\usepackage{colortbl}

\begin{document}

\title{{RSRefSeg: Referring Remote Sensing Image Segmentation with Foundation Models}
\thanks{$^\star$Corresponding author}}

\author{Keyan Chen, Jiafan Zhang, Chenyang Liu, Zhengxia Zou, Zhenwei Shi$^\star$\\
Beihang University}

\maketitle

\begin{abstract}

Referring remote sensing image segmentation is crucial for achieving fine-grained visual understanding through free-format textual input, enabling enhanced scene and object extraction in remote sensing applications. Current research primarily utilizes pre-trained language models to encode textual descriptions and align them with visual modalities, thereby facilitating the expression of relevant visual features. However, these approaches often struggle to establish robust alignments between fine-grained semantic concepts, leading to inconsistent representations across textual and visual information.
To address these limitations, we introduce a referring remote sensing image segmentation foundational model, RSRefSeg. RSRefSeg leverages CLIP for visual and textual encoding, employing both global and local textual semantics as filters to generate referring-related visual activation features in the latent space. These activated features then serve as input prompts for SAM, which refines the segmentation masks through its robust visual generalization capabilities.
Experimental results on the RRSIS-D dataset demonstrate that RSRefSeg outperforms existing methods, underscoring the effectiveness of foundational models in enhancing multimodal task comprehension. The code is available at \url{https://github.com/KyanChen/RSRefSeg}.

\end{abstract}


\section{Introduction}

Referring remote sensing image segmentation represents a frontier in vision-language multimodal understanding, as it aims to segment semantically meaningful regions or objects from remote sensing images using natural language descriptions \cite{yuan2024rrsis, liu2024rotated}. This approach transcends the limitations of traditional semantic segmentation methods in remote sensing by offering more flexible and granular segmentation capabilities. Such functionality is essential for comprehensive understanding of remote sensing scenes and objects. However, the task presents significant challenges, particularly in processing spatial and attribute relationships across both textual and visual modalities while maintaining precise semantic alignment in the latent space \cite{lei2024exploring}.

Recent advances in multimodal visual understanding have led to significant progress in referring remote sensing image segmentation. Early research relied on Convolutional Neural Networks (CNNs) and Recurrent Neural Networks (RNNs) for bimodal feature extraction, achieving segmentation through basic information fusion. The introduction of attention mechanisms subsequently enabled researchers to explore modal interaction and alignment in deep semantic space. Contemporary approaches predominantly employ pre-trained language models, such as BERT, to encode textual descriptions and guide the representation of relevant visual features based on extracted semantic information \cite{dong2024cross}. However, these methods face persistent challenges in achieving fine-grained semantic concept alignment and maintaining consistency in text-visual information expression. These challenges are particularly evident in capturing correlations between diverse ground objects and representing multi-scale remote sensing features, especially when dealing with small objects.

To address these challenges, this paper presents RSRefSeg, a foundation model for referring segmentation of remote sensing images. RSRefSeg achieves superior generalization and versatility by effectively leveraging coarse-grained text-visual semantics from CLIP \cite{radford2021learning} with refined mask representations from SAM \cite{kirillov2023segment}. Our research thoroughly investigates and addresses two key challenges: performance degradation during cross-domain transfer and the integration of general knowledge from multiple foundation models. Specifically, RSRefSeg leverages CLIP for visual and textual encoding, employing both global and local textual semantics as filters to generate referring-related visual activated features in the latent space. These activated features are then processed by the proposed AttnPrompter, which generates the necessary prompt inputs for SAM, ultimately leveraging SAM's robust segmentation capabilities to produce refined referring masks. To mitigate domain adaptation challenges, we introduce low-rank parameter-efficient fine-tuning in both CLIP and SAM backbones. The main contributions of this work are as follows,

1) We introduce a foundation model, RSRefSeg, which demonstrates exceptional generalization and versatility in the task of referring remote sensing image segmentation.

2) We investigate the potential of using text-visual coarse-grained aligned semantics from CLIP, as prompts for SAM to generate refined segmentation masks, addressing the challenges of performance degradation in cross-domain transfer and the difficulty of integrating and transferring general knowledge across multiple foundational models.

3) RSRefSeg outperforms contemporary methods on the RRSIS-D dataset, thereby highlighting the practical value of foundational models in understanding multimodal tasks within remote sensing.

\begin{figure*}[!thb]
\centering
\resizebox{0.9\linewidth}{!}{
\includegraphics[width=\linewidth]{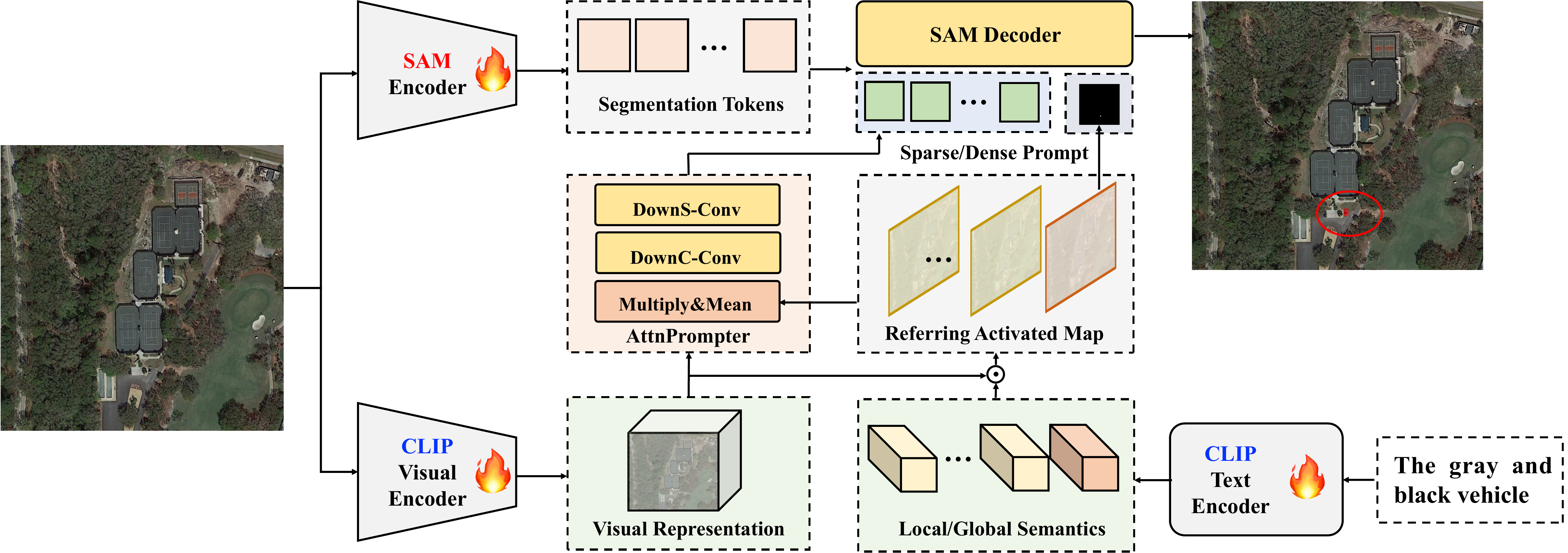}
}
\caption{The overview of the proposed RSRefSeg. The fire icon symbolizes that the model parameters are tuned.
}\label{fig:overview}
\vspace{-0.4cm}
\end{figure*}

\section{Methodology} \label{sec:method}
\vspace{-0.1cm}

\subsection{Overview} \label{sec:overview}
\vspace{-0.1cm}

RSRefSeg is developed by integrating the architectures of CLIP and SAM to achieve superior generalization and robust referring segmentation capabilities for remote sensing imagery. It comprises three principal components: a fine-tuned CLIP, an AttnPrompter, and a fine-tuned SAM. The fine-tuned CLIP extracts both global and local semantic embeddings from free-form textual referring and low-resolution visual features. The AttnPrompter processes CLIP's text and visual features to extract coarse visual activation features corresponding to the referred content, subsequently transforming them into prompt embeddings compatible with SAM. Finally, the fine-tuned SAM processes the original image using these prompt embeddings to generate the corresponding binary segmentation mask. The formulation is expressed as,
\begin{align}
\begin{split}
v,t &= \Phi_\text{ft-clip}(\mathcal{I}_1, \mathcal{T}) \\
p &= \Phi_\text{attn-prompter}(v,t) \\
\mathcal{M} &= \Phi_\text{ft-sam}(\mathcal{I}_2,p) \\
\end{split}
\end{align}
where $\mathcal{I}_1 \in \mathbb{R}^{H_1 \times W_1 \times 3}$ denotes the input remote sensing image, and $\mathcal{T} \in \mathbb{R}^{N}$ represents the corresponding textual description. These inputs are processed by the fine-tuned CLIP to generate a low-resolution visual feature map $v \in \mathbb{R}^{h_1 \times w_1 \times d_1}$ and text features $t \in \mathbb{R}^{(L+1) \times d_1}$, where $L+1$ indicates the number of tokens post-tokenization, including an additional end-of-sequence token (EOS). The generated prompt $p$ encompasses both sparse prompts (point or box embeddings) and dense prompts (coarse segmentation masks) produced by the attention prompter. The original image $\mathcal{I}_2 \in \mathbb{R}^{H_2 \times W_2 \times 3}$ and prompt $p$ are then processed by the fine-tuned SAM to produce the final segmentation mask $\mathcal{M} \in \mathbb{R}^{H_2 \times W_2}$. Notably, SAM's flexible prompt requirements mean that prompt $p$ need not be highly precise. Additionally, considering the characteristics of input from two different foundational models, we have $H_1 < H_2$, $W_1 < W_2$, meaning $\mathcal{I}_1$ is downsampled from $\mathcal{I}_2$.

\subsection{Fine-tuned CLIP} \label{sec:ft-clip}

\vspace{-0.1cm}

CLIP effectively encodes both free-text instructions and visual images, establishing a natural alignment between their extracted features. However, this alignment remains coarse-grained and lacks precision. Rather than pursuing high-precision semantic alignment, we leverage SAM's fine-grained segmentation capabilities to compensate for this limitation. Furthermore, while CLIP is primarily designed for general scene understanding, its performance deteriorates when applied directly to remote sensing domains due to distribution differences. To address this challenge, we introduce low-rank fine-tuning by introducing additional trainable parameters, as shown in the following equation,
\begin{align}
\begin{split}
W^\star = W + AB^T
\end{split}
\end{align}
where $W \in \mathbb{R}^{d \times d}$ represents the frozen parameter matrix in the foundational model, and $A \in \mathbb{R}^{d \times r}$ and $B \in \mathbb{R}^{d \times r}$ are newly introduced low-rank trainable parameter matrices, with $r \ll d$. These low-rank trainable parameters are incorporated into both image and text encoders.

While the original CLIP architecture produces sparse image and text representations for pre-training or classification, we modify it by removing the pooling layers to preserve the original image feature maps and hidden states for each text token. Notably, the end token consolidates semantic information for the entire sentence, with its corresponding embedding designated as the global semantic feature $t_\text{global} \in \mathbb{R}^{1 \times d_1}$. The feature embeddings of other words are classified as local semantic features $t_\text{local} \in \mathbb{R}^{L \times d_1}$, representing specific categories, locations, and other attributes. Thus, the complete feature representation is denoted as $t = [t_\text{local}, t_\text{global}]$.

\subsection{AttnPrompter} \label{sec:attn-prompter}
\vspace{-0.1cm}

SAM foundation model performs segmentation through various prompts, including points, boxes, and coarse segmentation masks. While SAM demonstrates powerful general segmentation capabilities, it lacks automated entity perception ability. This study does not aim to enhance this particular capability but rather employs SAM as an independent, decoupled mask generator controlled by referring semantic information. To integrate CLIP's referring semantic information as prompts into SAM, we propose AttnPrompter as a bridge between these two foundation models. AttnPrompter utilizes text semantics as filters to extract key visual features related to referring expressions and generates prompt embeddings (which can be interpreted as embeddings of points or boxes indicating segmentation targets) required by SAM through channel and spatial abstraction. The construction is formulated as follows,
\begin{align}
\begin{split}
v_\text{global-attn} &= v \cdot \sigma (v {t^T_\text{global}}) + v \\
v_\text{local-attn} &= \Phi_\text{mean}(v \cdot \sigma (v {t^T_\text{local}})) + v \\
v_\text{attn} &= \Phi_\text{cat}(v_\text{global-attn}, v_\text{local-attn}, v)
\end{split}
\end{align}
These formulas construct attended feature maps related to referring expressions, incorporating both spatial response features related to global text semantics and local related features. Here, $v \in \mathbb{R}^{h_1 \times w_1 \times d_1}$ represents the visual feature extracted by CLIP, and $\sigma (v {t^T_\text{global}})$ computes the activation map related to global semantics. Given $L$ local activation maps, we apply average pooling ($\Phi_\text{mean}$) and subsequently concatenate ($\Phi_\text{cat}$) the three components to obtain visual attended features ($v_\text{attn} \in \mathbb{R}^{h_1 \times w_1 \times d_1}$). The organization of $v_\text{attn}$ as prompt inputs required by SAM is defined as,
\begin{align}
\begin{split}
p_\text{sparse} &= \Phi_\text{conv-ds}(\Phi_\text{conv-dc}(v_\text{attn})) \\
p_\text{dense} &= \Phi_\text{resize}(v {t^T_\text{global}})
\end{split}
\end{align}
We have designed two types of prompts required by SAM: sparse prompts (analogous to point or box embeddings) and dense prompts (coarse segmentation masks). The sparse prompts $p_\text{sparse} \in \mathbb{R}^{M \times d_2}$ are derived from $v_\text{attn}$, where $M$ could be regarded as the number of prompt points. $\Phi_\text{conv-dc}$ denotes convolution with a $1 \times 1$ kernel size, reducing the channel dimension from $d_1$ to $d_2$. $\Phi_\text{conv-ds}$ represents multiple Conv-BN-GELU convolution blocks with a $3 \times 3$ kernel size and stride 2, employed for spatial dimension reduction to $M$. The dense prompts $p_\text{dense} \in \mathbb{R}^{h_2 \times w_2}$ comprise coarse-grained response masks obtained by filtering CLIP visual features with global semantics, followed by upsampling.

\subsection{Fine-tuned SAM} \label{sec:ft-sam}
\vspace{-0.1cm}

SAM processes the raw image $\mathcal{I}_2$ along with sparse/dense prompts from the prompt generator to produce the final referring segmentation mask through an encoding-decoding process. Since our task requires only a single segmentation output, we select the first generated mask from SAM's output as the final result. For comprehensive technical details, please refer to the SAM paper. To address the semantic distribution discrepancies in domain transfer, we introduce parameter fine-tuning in SAM's cumbersome encoder, following an approach similar to that outlined in Section \ref{sec:ft-clip}.

\section{Experiments}
\label{sec:experiment}

\subsection{Experimental Dataset and Settings}

We conducted experiments on the RRSIS-D dataset to validate the effectiveness of our proposed method. The dataset comprises 17,402 triplets, each containing an image, a mask, and a referring expression. The data was split into 12,181 training samples, 1,740 validation samples, and 3,481 test samples. RRSIS-D encompasses 20 distinct semantic categories, including aircraft, golf courses, highway service areas, baseball fields, and stadiums. All images are standardized to $800 \times 800$ pixels, with spatial resolutions varying from 0.5 to 30 meters.

\subsection{Evaluation Protocol and Metrics}

Following most previous research on referring segmentation, we assess performance using two primary metrics: generalized Intersection over Union (gIoU) and cumulative Intersection over Union (cIoU). The gIoU metric represents the mean of individual IoU scores across all images, whereas cIoU is computed as the ratio of cumulative intersection to cumulative union. We primarily rely on gIoU for our analysis, as cIoU tends to be skewed toward larger target areas and demonstrates greater statistical variance. Furthermore, we implement precision metrics (Pr@X) with thresholds ranging from 0.5 to 0.9 to quantify the percentage of test images meeting specific IoU criteria.

\subsection{Implementation Details}

\noindent \textbf{Architecture Details}: We employ SigLIP to encode the required prompts for SAM. SigLIP is an enhanced version of CLIP that employs a Sigmoid loss function during training. Specifically, we use the siglip-so400m-patch14-384 version. The prompter processes the visual attended feature maps from CLIP through two convolutional blocks for spatial downsampling, and then flattens them to serve as prompts for SAM. Additionally, the dimension for low-rank fine-tuning is set to $r = 16$. We conducted experiments using both the base and large versions of SAM, designated as RSRefSeg-b and RSRefSeg-l respectively, with ablation studies limited to the base version.

\begin{table*}[!thb]
\centering
\caption{
The results for referring image segmentation with various methods on the RRSIS-D dataset, with the best performance in bold.
}\label{tab:sota}
\resizebox{0.9\linewidth}{!}{
\begin{tabular}{c| c| *{5}{c} | c c}
\toprule
\textbf{Method} & \textbf{Publication} & \textbf{Pr@0.5} & \textbf{Pr@0.6} & \textbf{Pr@0.7} & \textbf{Pr@0.8} & \textbf{Pr@0.9} & \textbf{cIoU} & \textbf{gIoU} \\ 
\midrule
RRN \cite{li2018referring} & CVPR'2018 & 51.07 & 42.11 & 32.77 & 21.57 & 6.37 & 66.43 & 45.64 \\ 
CMSA \cite{ye2019cross} & CVPR'2019 & 55.32 & 46.45 & 37.43 & 25.39 & 8.15 & 69.39 & 48.54 \\ 
LSCM \cite{hui2020linguistic} & ECCV'2020 & 56.02 & 46.25 & 37.70 & 25.28 & 8.27 & 69.05 & 49.92 \\
CMPC \cite{huang2020referring} & CVPR'2020 & 55.83 & 47.40 & 36.94 & 25.45 & 9.19 & 69.22 & 49.24 \\ 
BRINet \cite{hu2020bi} & CVPR'2020 & 56.90 & 48.77 & 39.12 & 27.03 & 8.73 & 69.88 & 49.65 \\ 
CMPC+ \cite{liu2021cross} & TPAMI'2021 & 57.65 & 47.51 & 36.97 & 24.33 & 7.78 & 68.64 & 50.24 \\
LAVT \cite{yang2022lavt} & CVPR'2022 & 66.93 & 60.99 & 51.71 & 39.79 & 23.99 & 76.58 & 59.05 \\
CrossVLT \cite{cho2023cross} & TMM'2023 & 70.38 & 63.83 & 52.86 & 42.11 & 25.02 & 76.32 & 61.00 \\ 
LGCE \cite{yuan2024rrsis} & TGRS'2024 & 69.41 & 63.06 & 53.46 & 41.22 & 24.27 & 76.24 & 61.02 \\ 
RMISN \cite{liu2024rotated} & CVPR'2024 & 71.96 & 65.76 & 55.16 & 42.03 & 25.02 & 76.50 & 62.27 \\
FIANet \cite{lei2024exploring} & TGRS'2024 & 74.46 & 66.96 & 56.31 & 42.83 & 24.13 & 76.91 & 64.01 \\ 
\midrule
RSRefSeg-b (Ours) & & 72.82 & 67.35 & 56.66 & 45.86 & 26.78 & 76.05 & 63.68
\\
RSRefSeg-l (Ours) & & \cellcolor{gray!30}\textbf{74.49} & \cellcolor{gray!30}\textbf{ 68.33} & \cellcolor{gray!30}\textbf{58.73} & \cellcolor{gray!30}\textbf{48.50} & \cellcolor{gray!30}\textbf{30.80} & \cellcolor{gray!30}\textbf{77.24} & \cellcolor{gray!30}\textbf{64.67} 
\\
\bottomrule
\end{tabular}}
\end{table*}

\noindent \textbf{Training Details}: RSRefSeg employs binary cross-entropy loss for training. The model processes input images at different resolutions: $384 \times 384$ pixels for SigLIP and $1024 \times 1024$ pixels for SAM. The training procedure does not incorporate any additional data augmentations. In this configuration, the feature maps maintain dimensions of $h_1 = w_1 = 24$ and $h_2 = w_2 = 64$, with $M = 36$ in the sparse prompts. During training, both CLIP and SAM backbones remain frozen, with only the additional low-rank parameters being updated. The lightweight mask decoder of SAM is also included in the training process. For optimization, we employ AdamW with an initial learning rate of $1e-4$, with a cosine annealing learning rate scheduler and linear warmup strategy. The model trains for 200 epochs with a batch size of 32. The training is implemented on 8 NVIDIA A100 GPUs utilizing DeepSpeed ZERO 2 for distributed computing support.

\subsection{Comparison with the State-of-the-Art}

We evaluated the RSRefSeg with several state-of-the-art referring image segmentation methods, including RRN \cite{li2018referring}, CMSA \cite{ye2019cross}, LSCM \cite{hui2020linguistic}, CMPC \cite{huang2020referring}, BRINet \cite{hu2020bi}, CMPC+ \cite{liu2021cross}, LAVT \cite{yang2022lavt}, CrossVLT \cite{cho2023cross}, LGCE \cite{yuan2024rrsis}, RMISN \cite{liu2024rotated}, and FIANet \cite{lei2024exploring}. While LGCE, RMISN, and FIANet are specifically designed for remote sensing imagery, the other methods primarily target natural images. As shown in Table \ref{tab:sota}, RSRefSeg demonstrates superior performance, achieving cIoU and gIoU scores of 77.04 and 64.65, respectively. These results significantly surpass the performance of contemporary state-of-the-art methods, including LGCE, RMISN, and FIANet. The model's exceptional precision at higher IoU thresholds indicates its ability to generate highly accurate referring masks, thus validating the effectiveness of the foundation model in remote sensing referring image segmentation.

\subsection{Ablation Study} \label{sec:ablation}

\begin{table}[!hbt]
\centering
\caption{
Ablation results under different configurations in Sec. \ref{sec:ablation}.
}\label{tab:ablation}
\resizebox{0.98\linewidth}{!}{
\begin{tabular}{ c| c| *{5}{c} | c c}
\toprule
\textbf{ID} & \textbf{Setting}  & \textbf{Pr@0.5} & \textbf{Pr@0.6} & \textbf{Pr@0.7} & \textbf{Pr@0.8} & \textbf{Pr@0.9} & \textbf{cIoU} & \textbf{gIoU} 
\\ 
\midrule
\multirow{3}{*}{1} & 8 & 59.54 & 54.19 & 46.60 & 37.93 & 22.64 & 68.21 & 53.47
\\
&16 & \cellcolor{gray!30}\textbf{ 72.52} & \cellcolor{gray!30}\textbf{ 67.35} & \cellcolor{gray!30}\textbf{ 56.66} & \cellcolor{gray!30}\textbf{ 45.86} & \cellcolor{gray!30}\textbf{ 26.78} & \cellcolor{gray!30}\textbf{ 76.05} & \cellcolor{gray!30}\textbf{ 63.68}
\\
&32 & 70.22 & 64.82 & 55.86 & 45.00 & 25.91 & 75.28 & 62.64
\\
\midrule
\multirow{3}{*}{2}&zero &  57.98 & 52.76 & 44.59 & 35.23 & 21.38 & 64.92 & 51.62
\\
&half & 71.03 & 65.63 & 55.80 & 43.62 & 26.95 & 74.72 & 62.53
\\
&full & \cellcolor{gray!30}\textbf{ 72.52} & \cellcolor{gray!30}\textbf{ 67.35} & \cellcolor{gray!30}\textbf{ 56.66} & \cellcolor{gray!30}\textbf{ 45.86} & \cellcolor{gray!30}\textbf{ 26.78} & \cellcolor{gray!30}\textbf{ 76.05} & \cellcolor{gray!30}\textbf{ 63.68}
\\
\midrule
\multirow{3}{*}{3}&2 & 68.67 & 62.64 & 53.62 & 42.47 & 25.57 & 74.20 & 60.83
\\
&4 & \cellcolor{gray!30}\textbf{ 72.52} & \cellcolor{gray!30}\textbf{ 67.35} & \cellcolor{gray!30}\textbf{ 56.66} & \cellcolor{gray!30}\textbf{ 45.86} & \cellcolor{gray!30}\textbf{ 26.78} & \cellcolor{gray!30}\textbf{ 76.05} & \cellcolor{gray!30}\textbf{ 63.68}
\\
&8 & 66.37 & 61.32 & 52.06 & 41.37 & 24.25 & 72.70 & 59.17
\\
\midrule
\multirow{2}{*}{4}&w/o & 59.77 & 54.31 & 46.78 & 37.58 & 22.24 & 72.60 & 53.08
\\
& w/ & \cellcolor{gray!30}\textbf{ 72.52} & \cellcolor{gray!30}\textbf{ 67.35} & \cellcolor{gray!30}\textbf{ 56.66} & \cellcolor{gray!30}\textbf{ 45.86} & \cellcolor{gray!30}\textbf{ 26.78} & \cellcolor{gray!30}\textbf{ 76.05} & \cellcolor{gray!30}\textbf{ 63.68}
\\
\bottomrule
\end{tabular}}
\end{table}

To evaluate the impact of different components and fine-tuning strategies on segmentation performance, we conducted a series of ablation experiments using the previously described training settings. Table \ref{tab:ablation} presents the experimental results across four key aspects: 
1) We investigated the effect of varying the number of trainable parameters in the SAM encoder ($r=8$, 16, 32). The results demonstrate that an optimal number of trainable parameters is crucial for effective performance. Insufficient parameters hinder cross-domain knowledge transfer, while excessive parameters, despite facilitating rapid training convergence, lead to overfitting. 
2) With the CLIP encoder's rank fixed at 16, we experimented with fine-tuning different layers of the text encoder. The results reveal that fine-tuning the text encoder significantly enhances performance, challenging the conventional assumption that language models require no parameter adjustment due to domain invariance. This improvement may be attributed to two factors: first, referring expressions typically consist of phrases rather than complete sentences used in CLIP's original training; second, this task emphasizes discriminative semantics over general semantic understanding, suggesting the presence of domain-specific variations.
3) We tested different spatial downsampling rates (2, 4, and 8) in the prompter. The results indicate that an optimal prompt dimension is essential for superior performance. Excessive or insufficient prompts deviate from SAM's original training paradigm, resulting in suboptimal performance.
4) Finally, we evaluate the influence of dense prompts on overall performance. The results confirm that dense prompts provide superior priors for decoding, and their combination with sparse prompts yields optimal results. These comprehensive ablation studies offer valuable insights for downstream domain transfer learning in foundation models.

\section{Conclusion}

This paper addresses the limitations of current methods in fine-grained semantic alignment and text-visual consistency by leveraging foundational models' knowledge for referring remote sensing image segmentation. We introduce RSRefSeg, a foundation model comprising 1.2 billion parameters. The model's AttnPrompter architecture bridges CLIP and SAM foundation models by converting coarse-grained textual semantic activated visual features into prompt inputs for the SAM model, enabling the generation of precise referring masks. Our experimental evaluation on the RRSIS-D dataset demonstrates the effectiveness of RSRefSeg's components. The model's overall performance exceeds that of contemporary approaches, achieving state-of-the-art results and validating the efficacy of foundation models in understanding multimodal remote sensing tasks.

\small
\bibliographystyle{IEEEtranN}
\bibliography{references}

\end{document}